\documentclass[10pt,twocolumn,letterpaper]{article}

\usepackage{iccv}              

\usepackage{bm}

\definecolor{iccvblue}{rgb}{0.21,0.49,0.74}
\usepackage[pagebackref,breaklinks,colorlinks,allcolors=iccvblue]{hyperref}

\usepackage{multirow}
\usepackage{graphicx}
\usepackage{bbm}
\usepackage{float}
\usepackage{makecell}

\def\paperID{4} 
\def\confName{ICCV}
\def\confYear{2025}

\title{End-to-End Action Segmentation Transformer}

\author{Tieqiao Wang \hspace{20pt} 
Sinisa Todorovic\\
Oregon State University\\
{\tt\small \{wangtie, sinisa\}@oregonstate.edu}\\
{\small \url{https://github.com/tqosu/EAST}}
}

\begin{document}
\maketitle
\begin{abstract}
Most recent work on action segmentation relies on pre-computed frame features from models trained on other tasks and typically focuses on framewise encoding and labeling without explicitly modeling action segments. To overcome these limitations, we introduce the End-to-End Action Segmentation Transformer (EAST), which processes raw video frames directly -- eliminating the need for pre-extracted features and enabling true end-to-end training. Our contributions are as follows:  (1) a lightweight adapter design for effective fine-tuning of large backbones; (2) an efficient segmentation-by-detection framework for leveraging action proposals predicted over a coarsely downsampled video; and (3) a novel action-proposal-based data augmentation strategy. EAST achieves SOTA performance on standard benchmarks, including GTEA, 50Salads, Breakfast, and Assembly-101.
\end{abstract}    
\section{Introduction}
\label{sec:intro}

Action segmentation is a basic vision problem that involves labeling frames of an untrimmed video with their corresponding action classes. The problem poses many challenges, including the inherent ambiguity of action boundaries, and significant computational demands of processing long video sequences. 

Recent approaches typically address these challenges by using pre-computed frame features, e.g., I3D \cite{I3D} or TSM \cite{lin2019tsm}. Such features are known to be suboptimal \cite{ding2023temporal}, because they have been previously extracted by other methods trained on vision tasks that are different from action segmentation (e.g., action recognition). Furthermore, due to memory and computational constraints, most  approaches focus on framewise representations \cite{AbuFarha2019MSTCNMT, yi2021asformer, liu2023diffusion} that lack explicit modeling of action instances  (with few exceptions \cite{lu2024fact, wang2024efficient} that  increase complexity). Consequently, they neglect the bottom-up/top-down integration of frame and action representations, which was once essential in traditional frameworks \cite{BrendelICCV11,PeiICCV11,SiPYZ11, AmerCVPR13,HuangCVPR20}. Finally, as datasets for action segmentation are significantly smaller than for other tasks (e.g., action recognition), previous work resorts to data augmentation and self-supervised learning \cite{aziere2023markov,Li_2021_CVPR, MisraZH16}. 
However, these methods typically augment only local frame features, and do not augment action instances.  
\begin{figure*}
    \centering
    \includegraphics[width=\linewidth]{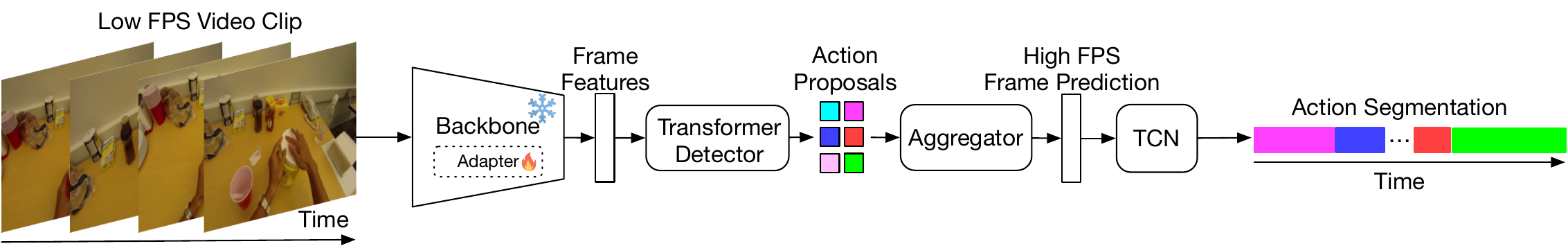}
    \caption{EAST consists of a frozen backbone with trainable adapters for efficient feature adaptation, a transformer-based detector for regressing action boundaries over coarsely sampled frames (low FPS), an aggregator for combining the action proposals to predict a class distribution of every frame at the original frame-rate (high FPS), and a refinement module to perform final framewise classification.}
    \label{fig:overview}
    \vspace{-10pt}
\end{figure*}

To address these limitations, we propose the End-to-End Action Segmentation Transformer (EAST). As shown in Fig.~\ref{fig:overview}, EAST consists of the following modules: (a) a large backbone, (b) a detector that predicts action proposals over coarsely sampled frames; (c) an aggregator that combines the proposals to infer class distributions for all frames at the original (unsampled) frame rate; and (d) a refinement module for predicting the final framewise labels. With EAST, we make the following three key contributions. 

First, we enable efficient end-to-end training of EAST by introducing lightweight Contract-Expand Adapters (CEA) into a large backbone network. CEA reduces complexity by compressing and expanding features around depth-wise convolutions. This allows efficient fine-tuning of the large backbone to extract multiscale features directly from raw RGB frames, ensuring they are optimized for action segmentation rather than for unrelated vision tasks. While recent methods such as Bridge-Prompt \cite{2022_CVPR_Li} and FACT \cite{lu2024fact} also claim end-to-end training from RGB inputs, their training strategy consists of disjoint stages -- first training the backbone, then freezing features to train the rest of their segmentation model. In contrast, EAST is trained truly end-to-end, without freezing features at any stage.

Second, unlike most recent work that focuses on framewise labeling without explicitly modeling the temporal extent of action instances, EAST performs action segmentation by detection. Specifically, EAST detects action proposals over a downsampled input by regressing action boundaries over the sampled frames. The proposals are then integrated and refined for the final segmentation. This provides two key advantages: improved efficiency by detecting action proposals on coarsely sampled frames (e.g., 1–6 fps), and enhanced framewise classification by explicitly considering the context of detected action segments.  It is worth noting that temporal downsampling does not affect the ground truth or evaluation, as our boundary regression is specified relative to timestamps within the video. These boundary detections are mapped to the original full frame rate, and serve as useful constraints for final action segmentation over all frames.

Third, we introduce a novel proposal-based data augmentation method to enhance EAST training. In training, our experiments show that the high-confidence action proposals generally align well with the ground truth, creating an easier learning signal for the subsequent proposal aggregation and refinement stages. To encourage learning from more challenging cases, we augment training by selectively removing high-confidence proposals before passing the remaining proposals to the subsequent EAST modules.

With these contributions, EAST achieves state-of-the-art performance on standard benchmarks, including the GTEA, 50Salads, Breakfast, and Assembly-101 datasets. Our approach outperforms existing methods across all metrics.

\section{Related Work}
\label{sec:related-works}
This section reviews closely related work.

{\bf Efficient Training}.
While end-to-end training offers known advantages, memory and computational constraints often make it impractical. Parameter-efficient fine-tuning (PEFT) methods, such as adapters \cite{houlsby2019parameter}, LoRA \cite{hu2021lora}, and prefix-tuning \cite{li2021prefix}, address these limitations by reducing the number of trainable parameters. However, PEFT's potential for video understanding, particularly action segmentation, remains largely unexplored. AdaTAD \cite{liu2024end} introduces Temporal-Informative Adapters (TIA) for action detection, using depth-wise convolutions (DWConv) \cite{howard2017mobilenets} to enhance temporal reasoning. While TIA improves performance over standard adapters \cite{houlsby2019parameter}, it also increases complexity and slows convergence. To address this, we propose the Contract-Expand Adapter (CEA), designed specifically for action segmentation. CEA applies feature compression and expansion around the DWConv. This reduces the computational load within the adapter, achieving both the performance gains of TIA and the benefits of standard adapters -- lower complexity and faster convergence.

{\bf Temporal Action Segmentation} has been tackled with multi-stage framewise networks like MS-TCN \cite{AbuFarha2019MSTCNMT}, ASFormer \cite{yi2021asformer}, and DiffAct \cite{liu2023diffusion}. However, these lack explicit action instance representations and require post-processing, hindering end-to-end training. More recent approaches (UVAST \cite{uvast2022ECCV}, FACT \cite{lu2024fact}, BaFormer \cite{wang2024efficient}) model actions using query tokens alongside frame tokens, but at the cost of significantly increased complexity. 

Importantly, most recent methods operate on all frames, at the input frame rate, and do not use action boundaries to constrain framewise labeling \cite{AbuFarha2019MSTCNMT, yi2021asformer}. This limits their ability to handle downsampled videos, a critical requirement for efficient end-to-end training with long videos. In contrast, we perform efficient action-boundary regression on multi-scale frame features, followed by the integration of the action proposals "top-down" for final framewise classification. This enables competitive performance of EAST even with downsampled input, significantly reducing model and computational complexity, compared to methods requiring full, unsampled video sequences \cite{AbuFarha2019MSTCNMT, yi2021asformer}.

{\bf Test-Time Post-Processing and Data Augmentation.} To improve performance, some approaches resort to post-processing, such as, e.g., Viterbi decoding \cite{uvast2022ECCV}. However, Viterbi decoding is computationally expensive and incompatible with end-to-end training. To enable robust training on relatively small action segmentation datasets, prior work uses data augmentation \cite{liu2023diffusion, aziere2023markov}, which are either  simplistic -- e.g., feature masking \cite{liu2023diffusion} -- or overly complex -- e.g., reinforcement learning based sequence generation \cite{aziere2023markov}. The latter would be difficult to optimize within an end-to-end framework. 
In contrast, we introduce a new data augmentation method that manipulates action proposals to enforce EAST training under higher uncertainty conditions,  seamlessly integrating within our end-to-end training. To our knowledge, this is the first work to apply proposal-based data augmentation for action segmentation.

\section{Specification of EAST}

\label{sec:proposed}

EAST consists of a backbone, detector, integrator, and refinement module, as shown in Fig.~\ref{fig:overview}. Given an untrimmed RGB video,  \(\mathbf{V} \in \mathbb{R}^{T \times H \times W \times 3}\), as input, 
the backbone takes a downsampled sequence,  \(\mathbf{V}' \in \mathbb{R}^{T' \times H \times W \times 3}\), where \(H\) and \(W\) are the frame height and width, and \(T'\) is the number of coarsely sampled frames, \(T'\ll T\). Frames are uniformly sampled at an empirically optimized rate to facilitate efficient end-to-end training. The backbone output is passed to the detector to predict: (i) Initial frame labels, \(\hat{\mathcal{Y}}_1 = \{(t_i, \hat{y}_i)\}_{i=1}^{T'}\), where \(t_i\) is the timestamp of frame $i$ in the original (unsampled) video  \(\mathbf{V}\),  and \(\hat{y}_i\) is its predicted class; and (ii) Action proposals, \(\hat{\mathcal{S}} = \{(\hat{t}_{n}^s, \hat{t}_{n}^e, \bm{\pi}_n)\}_{n=1}^{N}\), where \(N\) is the number of action proposals, \(\hat{t}_{n}^s\) and \(\hat{t}_{n}^e\) denote the predicted start and end timestamps  in \(\mathbf{V}\) of  \(n\)th action proposal, and \(\bm{\pi}_n\) is the predicted class distribution, \(\bm{\pi}_n=\{\pi_n(a):a\in\mathcal{A}\}\), over the set of action classes, $\mathcal{A}$, including a background class. For each sampled frame in \(\mathbf{V}'\), the detector regresses timestamps  of action boundaries in the unsampled \(\mathbf{V}\), rather than their frame indices. This enables the use of variable downsampling frame rates based on available memory and computational resources. 

EAST's integrator takes the action proposals in \(\hat{\mathcal{S}}\) as input, and combines them to predict a class distribution of every frame of the unsampled \(\mathbf{V}\). These predictions are then progressively refined through multiple stages of the standard Temporal Convolutional Network (TCN) \cite{AbuFarha2019MSTCNMT} for final framewise classification, \(\hat{\mathcal{Y}}_2\), over all $T$ frames.

Our end-to-end training  uses the proposed new data augmentation method, where \(\hat{\mathcal{S}}\) is corrupted by randomly removing a subset of the most confident proposals, before producing \(\hat{\mathcal{Y}}_2\). In the following, we provide a more detailed specification of EAST.

\subsection{Contract-Expand Adapter}

\begin{figure}[h]
	\centering
	\includegraphics[width=0.8\linewidth]{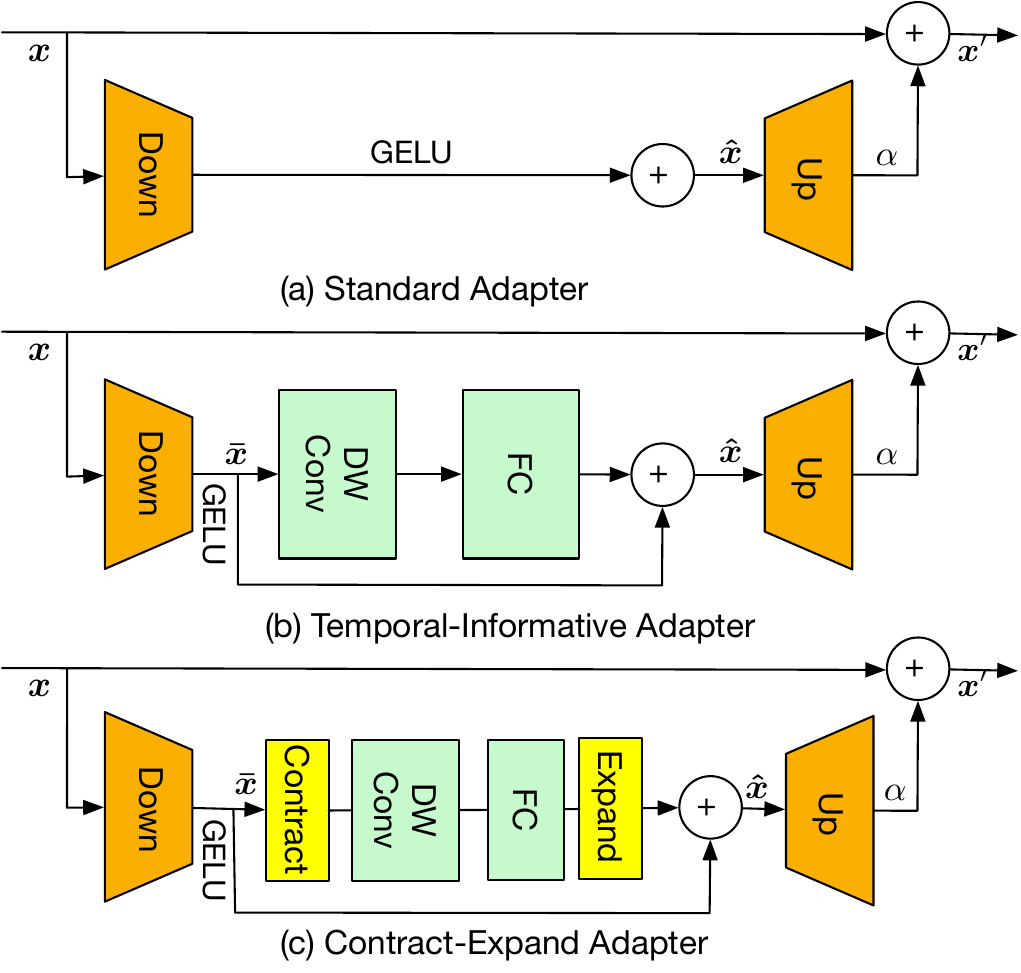}
	\caption{Adapters for efficient fine-tuning of a foundation backbone are typically inserted between its layers. (a) Standard Adapter \cite{houlsby2019parameter} (orange). (b) Temporal Interaction Adapter (TIA)  \cite{liu2024end} (green). (c) Our Contract-Expand Adapter (CEA) (yellow). CEA consists of temporal depth-wise convolutions \cite{howard2017mobilenets}, and parameter-free contract/expand layers.}
	\label{fig:adapter_arch}
	\vspace{-15pt}
\end{figure}

In training, EAST fine-tunes a pre-trained video foundation model on a given action segmentation dataset. As the backbone, we use ViT-G \cite{zhai2022scaling}, pre-trained with VideoMAEv2 \cite{wang2023videomae} on related vision tasks. To facilitate efficient end-to-end training within memory and computational constraints, we design a lightweight Contract-Expand Adapter (CEA), and integrate it into ViT-G. Building on the recent approaches to feature adaptation \cite{houlsby2019parameter, liu2024end}, we insert CEA between the backbone's layers. As shown in Fig.~\ref{fig:adapter_arch}, CAE  adapts the features $\bm{x}$ of the previous layer with a residual, which results in the adapted features $\bm{x}'$ that are further passed to the following layer. 

Fig.~\ref{fig:adapter_arch} illustrates key differences of CEA from previous approaches. The Standard Adapter \cite{houlsby2019parameter} consists of  down-projection and up-projection layers with a non-linear activation. However, the Standard Adapter does not explicitly model temporal context, making it unsuitable for action segmentation. The Temporal Interaction Adapter (TIA) \cite{liu2024end} incorporates temporal depth-wise convolutional layers (DWConv) to aggregate temporal context. TIA first reshapes a given input feature of shape $(B, C, T, H, W)$ to $(B\times H\times W, C, T)$, and then applies the same DWConv independently to each spatial location $(h, w)\in H\times W$. This results in a high computational cost, which would make TIA very challenging to incorporate in our end-to-end training. To meet our memory and computational constraints, we adopt a simpler adapter design as follows.

Our key idea is to use spatial average pooling directly within the adapter to reduce the number of spatial locations  $|H\times W|$ that share the same DWConv. 
This is based on the hypothesis that, during backbone fine-tuning, spatial context is less crucial for feature adaptation than temporal context. Our spatial average pooling significantly reduces data flow and computational complexity compared to TIA. After applying DWConv to a few pooled spatial locations $(h, w)$, the resulting features are then appropriately copied to the other $H\times W$ locations, spatially upscaling the enhanced features before they are passed to the next backbone layer.

As our results show, not only does our contract-and-expand strategy reduce GFLOPs, but it also improves both convergence speed and overall performance. Placing the spatial pooling outside the down- and up-projection layers of the adapter degrades both performance and convergence speed, highlighting the importance of its integration within the core structure of our adapter.

CEA's operations include the following:
\begin{equation}
\begin{aligned}
 \bm{\bar x} &= \bm {\sigma} ({\bm W_{\rm down}^{\rm \top}} \cdot \bm{x} ) \\
 \bm{\bar x}_c &=  \text{contract}_{HW}(\bm{\bar x}) \\
  \bm{\hat x}_m &= {\bm W_{\rm mid}^{\rm \top}} \cdot \operatorname{\textbf{DWConv}}_k(\bm{\bar x}_c) \\
  \bm{\bar x}_e &=  \text{expand}_{HW}(\bm{\hat x}_m) \\
 \bm{\hat x} &=   \bm{\bar x}_e + \bm{\bar x} \\
 \bm{x'}  &= \alpha \cdot {\bm W_{\rm up}^{\rm \top}} \cdot \bm{\hat x}  + \bm {x}
\end{aligned}
\label{eq:our_adapter}
\end{equation}
where \(\bm{x}\) and \(\bm{x'}\) are the input and output features, and \(\bm{\bar{x}}\) and \(\bm{\hat{x}}\) are intermediate features, as illustrated in Fig.~\ref{fig:adapter_arch}.
\(\bm{W}_{\rm down}\) and \(\bm{W}_{\rm up}\) are projection weights, \(\bm{W}_{\rm mid}\) are weights of an intermediate fully-connected layer, \(\operatorname{\textbf{DWConv}}_k\) is the depth-wise convolution,  \(\alpha\) is a learnable scalar, and  \(\bm {\sigma}(\cdot)\) is the GELU activation function \cite{hendrycks2016gaussian}.

During fine-tuning, only the CEA modules inserted between the backbone layers are trained, while the backbone remains frozen. With a temporal kernel size of 3 and a channel downsampling ratio of 4, the CEA comprises just 4.7\% of the backbone's parameters. The CEA's GFLOPs are almost identical to the Standard Adapter. Compared to the Standard Adapter, CEA increases GFLOPs by an additional 0.04, while TIA increases it by an additional 5.8 GFLOPs.

\subsection{Low-Frame-Rate Action Detection}
\label{sec:detector}
While accounting for temporal context is widely recognized as beneficial for action segmentation, memory and time complexity constraints often limit the video length that can be analyzed. Therefore, temporal downsampling seems like a critical strategy to manage computational resources, especially for our end-to-end training. However, state-of-the-art (SOTA) action segmentation models typically struggle to maintain the high accuracy achieved at high frame rates (FPS) when applied to low-FPS input. 

To enable efficient end-to-end training with temporal downsampling, we adopt a segmentation-by-detection framework, departing from SOTA approaches. Action proposals are predicted on coarsely sampled frames and then integrated at the original high frame rate for framewise classification. Inspired by anchor-free detectors (e.g., FCOS \cite{tian2019fcos} and ActionFormer \cite{zhang2022actionformer}), we treat each sampled frame as a query for its corresponding action proposal.

This generates high-quality action proposals from low-FPS input. Compared to SOTA methods that rely on separate frame and action branches \cite{lu2024fact} or learnable queries \cite{uvast2022ECCV}, our approach significantly simplifies training by directly predicting action instances from sampled frames.

We first feed the backbone output  \(\mathcal{X}\) to a transformer encoder, which produces a multiscale feature pyramid \(\mathcal{Z} = \{(\mathbf{z}_i^1, \dots, \mathbf{z}_i^L)\}_{i=1}^{T'}\), capturing long-range temporal dependencies. This encoder consists of a shallow convolutional projection followed by a Transformer network with a multi-head self-attention, which operates at varying temporal scales by downsampling with strided depthwise 1D convolutions. The transformer encoder output is passed to a convolutional decoder with  classification and regression heads. The
classification head uses a 1D convolution across the $L$ pyramid levels in \(\mathcal{Z}\) to predict  the class distribution of every frame, $\bm{ \mathcal{\pi}}_i=\{ \pi_i(a): a\in\mathcal{A}\}$, $i=\{1,\dots,T'\}$, where $\mathcal{A}$ is the set of action classes including a background class, $\pi_i(a)\in[0,1]$, and $\sum_{a\in\mathcal{A}}\pi_i(a)=1$. Simultaneously, for every frame $i$, the regression head convolves \(\mathcal{Z}\) across the levels to predict time offsets $\hat{d}^s_i$ and $\hat{d}^e_i$ to the start and end  timestamps of the action instances to which $i$th frame belongs. In this way, every frame $i$ generates the corresponding action proposal with the start and end timestamps estimated as  $\hat{t}_i^s = t_i - \hat{d}^s_i$ and $\hat{t}_i^e = t_i + \hat{d}^e_i$.

In summary, the detector of EAST performs structured prediction \(\mathcal{X} \to \{(\bm{\pi}_i,\hat{t}^s_i,\hat{t}^e_i) \}_{i=1}^{T'}\), which is mapped to the initial frame labels $\hat{\mathcal{Y}}_1$ as  $\hat{y}_i = \text{argmax}_{a\in\mathcal{A}} \; \pi_i(a)$, and the set of $T'$ action proposals 
\(\hat{\mathcal{S}} = \{(\hat{t}_{n}^s, \hat{t}_{n}^e, \bm{\pi}_n)\}_{n=1}^{N}\),  $N=T'$. 

\subsection{High-Frame-Rate Aggregation and Refinement}
\label{sec:integrator}

After generating action proposals \(\hat{\mathcal{S}}\) from the downsampled $\mathbf{V}'$, they are combined to estimate the class distributions, $\bm{p}_i=\{p_i(a): a{\in} \mathcal{A}\}$, of all frames  in the unsampled $\mathbf{V}$. To this end, each frame $i=1,\dots,T$ of $\mathbf{V}$ aggregates the class distributions \(\{\bm{\pi}_n \}\) of all proposals  whose temporal intervals cover the timestamp of $i$th frame, $\hat{t}^s_n \le t_i \le \hat{t}^e_n$:
\begin{align}
    p_i(a) &\propto \sum_{n=1}^{N}  \pi_n(a) \cdot \mathbbm{1}(\hat{t}^s_n \le t_i \le \hat{t}^e_n),
    \label{eq:aggregation}
\end{align}

 where ``$\propto$'' denotes proportionality up to a normalizing constant, $\mathbbm{1}(\cdot)$ is a binary indicator, and  \(\sum_{a {\in} \mathcal{A}} p_i(a) {=} 1\).

The aggregated class distributions of all frames, \(\{\bm{p}_i\}_{i=1}^T\), are  passed to a 3-stage temporal convolutional network (TCN) \cite{AbuFarha2019MSTCNMT} for final frame classification \(\hat{\mathcal{Y}}_2\). Similar refinement strategies are used in MS-TCN \cite{AbuFarha2019MSTCNMT} and ASFormer \cite{yi2021asformer}. The aggregation of action proposals in \eqref{eq:aggregation} helps improve temporal smoothness of frame labels in \(\hat{\mathcal{Y}}_2\) relative to the initial prediction \(\hat{\mathcal{Y}}_1\).

\subsection{Training Loss Functions}
\label{sec:training}

In training, predictions $\hat{\mathcal{Y}}_1$ and $\hat{\mathcal{S}}$ over sampled frames in $\bm{V}'$, $i=1,\dots,T'$, incur loss, $\mathcal{L}$, defined as
\begin{equation}
\mathcal{L} = \frac{1}{T_+}\sum_{i=1}^{T'} \mathcal{L}_{\text{c}}(y_i,\hat{y}_i) + \lambda_{\text{r}} \mathbbm{1}(\hat{y}_i) \mathcal{L}_{\text{r}}([t_{n(i)}^s, t_{n(i)}^e],[\hat{t}_{i}^s, \hat{t}_{i}^e]),
\end{equation}
where $\mathcal{L}_{\text{c}}$ denotes the focal loss \cite{8417976};  $y_i$ is the ground-truth class; $\lambda_{\text{r}}$ is a weighting hyperparameter; $\mathbbm{1}(\cdot)$ is a binary indicator that equals 0 if $i$th frame is classified as background, and 1, otherwise; $T_+$ is the total number of sampled frames classified as an action \(
T_+ = \sum_{i=1}^{T'} \mathbbm{1}(\hat{y}_i)
\); $\mathcal{L}_{\text{r}}$ is the  DIoU loss \cite{Zheng_Wang_Liu_Li_Ye_Ren_2020} for regression; $[t_{n(i)}^s, t_{n(i)}^e]$ denotes the time interval of $n$th ground-truth action instance closest to the predicted interval $[\hat{t}_{i}^s, \hat{t}_{i}^e]$ of $i$th action proposal. Note that the regression loss is applied only to action proposals classified as an action excluding background.

The final frame classification $\hat{\mathcal{Y}}_2$ is supervised with the cross entropy loss and smoothness loss, as in \cite{AbuFarha2019MSTCNMT}.

\subsection{Proposal-Based Data Augmentation}
\label{sec:augmentation}

This section introduces a new proposal-based data augmentation, which is seamlessly integrated into our end-to-end training. It enforces high uncertainty in the input to the aggregation and TCN modules, reflecting the likely conditions during testing. Integrating existing data augmentation methods that manipulate or generate frame sequences into end-to-end training is challenging due to memory and complexity constraints. In contrast, our method is both efficient and effective, as it operates on significantly fewer proposals than frames in the video.

Our method randomly removes $K< A$ out of the top $A$ most confident action proposals from $\hat{\mathcal{S}}$, resulting in $\hat{\mathcal{S}}'$. In our experiments, we choose $A=30$, because videos of the existing action-segmentation datasets typically have a maximum of 30 action instances. Confidence of a proposal, $\kappa_n$, is estimated as its maximum class score $\kappa_n=\max_{a\in \mathcal{A}}\pi_n(a)$. 
This reduces the average confidence of the remaining proposals in $\hat{\mathcal{S}}'$  passed to the aggregation module, where they ``compete'' under increased uncertainty to assign their class distributions to frames in \eqref{eq:aggregation}. It is worth noting that due to the random removals,  $\hat{\mathcal{S}}'$ may still include some of the top-scoring proposals from $\hat{\mathcal{S}}$,  which facilitates prediction of $\hat{\mathcal{Y}}_2$. Multiple random proposal removals are applied to generate multiple versions of $\hat{\mathcal{S}}'$, and thus perform data augmentation.

\section{Results}
\label{sec:experiments}

{\bf Datasets.} For evaluation, we use  the GTEA \cite{GTEA}, 50Salads \cite{50Salads},  Breakfast \cite{Breakfast}, and Assembly101 \cite{Assembly101} datasets. 
\begin{itemize}
    \item \textbf{GTEA} \cite{GTEA} consists of 28 egocentric videos, annotated with 11 action classes. The videos span approximately 1 minute and include around 19 action instances.
    \item  \textbf{50Salads}~\cite{50Salads} is a collection of 50 top-view videos of salad preparation with 17 action classes. The average length of these videos is 6 minutes, with approximately 20 action instances per video. 
    \item \textbf{Breakfast}~\cite{Breakfast} consists of 1712 videos showing 48 breakfast preparation actions from a third-person perspective. The videos have an average length of 2 minutes, but they may significantly vary in duration.
    \item \textbf{Assembly101}~\cite{Assembly101} consists of 4321 videos and 202 action classes based on 11 verbs and 61 objects. The dataset features people assembling and disassembling 101 toys. The videos have on average 24 action instances over a duration of 7.1 minutes. 
\end{itemize} 

Consistent with previous work, we perform five-fold cross-validation on 50Salads, and four-fold cross-validation on GTEA and Breakfast, using the standard splits \cite{uvast2022ECCV,2022_NeurIPS_Xu,yi2021asformer,2020_PAMI_Li}. For Assembly101, we use the official training and validation splits specified in \cite{Assembly101}.

{\bf Metrics.} As in SOTA action segmentation approaches,
we use framewise classification accuracy (Acc), edit score (Edit), and F1-scores (F1@$\{10, 25, 50\}$) at overlap thresholds of 10\%, 25\%, and 50\% \cite{lea2016temporal}. Edit score measures similarity between the predicted and ground-truth action sequences. F1-scores evaluate localization of action instances. We also report the average precision at Intersection over Union (IoU) thresholds of $\{0.3, 0.4, 0.5, 0.6, 0.7\}$ and the mean average precision (mAP) of our action proposals predicted by EAST's detector.

{\bf Implementation Details.} EAST is implemented using PyTorch 2.0.1 and MMAction2 \cite{2020mmaction2} on H100 GPUs. EAST consists of a backbone, detector, aggregator and TCN. The backbone is VideoMAEv2 \cite{wang2023videomae} with ViT-G \cite{zhai2022scaling}, pre-trained as in \cite{wang2023videomae}. Parameters of the backbone are frozen to their pre-trained values, and fine-tuned with our CEA adapters 
placed between ViT-G blocks of VideoMAEv2, with the adapter's learning rate set to 2e-4. The adapter's projection layer weights and \(\alpha\) are initialized to 0 and 1, respectively. 

During training, video clips are randomly cropped to 768 frames. The frame sampling rate (FPS) is treated as an empirically optimized hyperparameter tested in Tab.~\ref{tab:fps}. We use the following FPS: 3 for GTEA, 1 for 50salads, 3 for Breakfast, 6 for Assembly101. Backbone frame features, $\mathcal{X}$, are extracted using non-overlapping temporal windows of size 16 (stride = 16 frames) and a spatial resolution of 160x160. For inference on videos exceeding 768 frames, a 0.25 overlap  sliding window approach is employed. Predicted action boundary timestamps are regressed directly in seconds. These are then multiplied by the FPS to generate high-FPS frame-wise predictions, ensuring consistency with prior work for comparison. EAST is trained in two stages for stable convergence.  First, the backbone and detector are trained end-to-end for 300 epochs on GTEA, 150 epochs on 50Salads, 30 epochs on Breakfast, and 15 epochs on Assembly101. Subsequently, the entire EAST is trained for 50 epochs using our  data augmentation method. As in \cite{liu2023diffusion,lu2024fact}, model selection is performed on the validation set based on the average  metrics.

\subsection{Features and Training Efficiency}
In this section, we compare EAST with SOTA on the two largest datasets Breakfast and Assembly101 to evaluate: frame feature representation and training efficiency. As SOTA representatives, for comparison, we choose FACT \cite{lu2024fact}, LTContext \cite{bahrami2023much}, ASFormer \cite{yi2021asformer}, and MSTCN \cite{AbuFarha2019MSTCNMT}.

The SOTA methods use precomputed I3D features \cite{I3D} for Breakfast and TSM features \cite{lin2019tsm} for Assembly101.  To ensure consistency in the frame features across the SOTA methods and EAST, we extracted MAEv2 frame features for all the methods, including ours, using the ViT-G backbone pretrained with VideoMAEv2 \cite{wang2023videomae}. Tab.~\ref{tab:01_maev2_breakfast} shows that the SOTA methods  improve performance on Breakfast when using precomputed MAEv2 features, while LTContext with MAEv2 features fails to do so on Assembly101. On both datasets, end-to-end trained EAST using the same pretrained ViT-G achieves  superior performance.

\begin{table}
	\centering
\resizebox{0.9\columnwidth}{!}{
	\begin{tabular}{c|c|ccccc}
		\hline
		Method&Feature&\multicolumn{3}{c}{F1@\{10,25,50\}}&Edit&Acc\cr
			\hline
				MSTCN \cite{AbuFarha2019MSTCNMT}&I3D&52.6&	48.1&37.9&	61.7	&66.3	\cr	
			('CVPR19) &MAEv2*&59.9 & 55.1 & 43.9 & 65.0 & 68.5	\cr	
			\hline
			ASFormer \cite{yi2021asformer}&	I3D&	76.0&70.6&57.4&		75.0&73.5	\cr	
			('BMCV21) &MAEv2*&78.7 & 73.6 & 60.8 & 76.8 & 75.0\cr
			\hline
			{LTContext}  \cite{bahrami2023much}&I3D&77.6 &72.6&60.1&77.0&74.2\cr
				('ICCV23)			&MAEv2*&80.6 & 75.7 & 64.1 & 75.2 & 76.6	\cr	
			\hline
\multirow{3}{*}{\makecell{FACT \cite{lu2024fact}\\('CVPR24)}}&I3D &81.4& 76.5& 66.2& 79.7 &76.2 \cr
       &I3D* &78.4 & 73.3 & 62.2 & 77.4 & 75.5 \cr
	&MAEv2*&80.6 & 75.9 & 65.3 & 78.9 & 77.2\cr	
\hline
EAST&MAEv2&\textbf{85.6} & \textbf{81.5} & \textbf{71.6} & \textbf{83.5} & \textbf{82.2} \cr

		\hline
	\end{tabular}
    }
\caption{Impact of feature representation on Breakfast. *: results from official code. SOTA methods use precomputed I3D features on Breakfast. To ensure a consistent feature representation, we compare SOTA and EAST using MAEv2 frame features extracted by the ViT-G backbone pretrained with VideoMAEv2 \cite{wang2023videomae} across all the methods. Results are averaged across four standard splits. }
\label{tab:01_maev2_breakfast}
\end{table}

\begin{table}
	\centering
\resizebox{0.9\columnwidth}{!}{
	\begin{tabular}{c|c|ccccc}
		\hline
		Method&Feature&\multicolumn{3}{c}{F1@\{10,25,50\}}&Edit&Acc\cr
			\hline
			{LTContext}  \cite{bahrami2023much}&TSM&33.9 & 30.0& 22.6 & 30.4 & 41.2\cr
				('ICCV23)			&MAEv2&31.3&27.9&21.1&27.8&40.3	\cr	
			\hline
EAST&MAEv2&{\bf 42.3} & {\bf39.4} & {\bf32.8} & {\bf39.9} & {\bf48.4} \cr

		\hline
	\end{tabular}
    }
\caption{Impact of feature representation on Assembly101. LTContext \cite{bahrami2023much} uses precomputed TSM features on Assembly101. For fair comparison, we evaluate LTContext and EAST when both methods use MAEv2 frame features extracted by the ViT-G backbone pretrained with VideoMAEv2 \cite{wang2023videomae}.
}
\label{tab:01_maev2_assembly101}
\end{table}

\begin{table}
		\centering
        \resizebox{\columnwidth}{!}{
		\begin{tabular}{c|ccccc|c}
			\hline
			Methods &MSTCN++&ASFormer& LTContext & DiffAct & FACT&EAST\cr
            	\hline
GFLOPS &4.5 & 7.6 & 7.8 & 63.0 & 5.5&9.2\cr
              
			\hline
		\end{tabular}
        }
  \caption{Computational cost in GFLOPS, measured on a 5-minute video using the setup from \cite{lu2024fact}.}
  \label{tab:parameter}
\end{table}

Tab.~\ref{tab:abl_time} compares per-epoch time, total training time, and GPU memory usage of SOTA methods and EAST on Breakfast using NVIDIA H100s. Note that the SOTA methods in  Tab.~\ref{tab:abl_time}  rely on pre-computed features and in their reports exclude feature extraction time and memory usage. Therefore, the comparison of training efficiency is unfavorable to EAST, whose reported numbers in  Tab.~\ref{tab:abl_time} account for the full pipeline, including training the large backbone. Tab.~\ref{tab:abl_time} shows EAST achieves total training time comparable to LTContext -- the fastest feature-based Transformer -- despite processing RGB frames.

For a fair comparison with prior work that does not support direct RGB input and relies on pre-computed features, we account for feature extraction costs in   Tab.~\ref{tab:abl_fea_extract} in the supplement. As Tab.~\ref{tab:abl_fea_extract} demonstrates, SOTA methods require significantly more resources -- e.g., FACT needs 300.3 hours for I3D feature extraction plus 122.3 hours for model training (total 422.6 hours), making it 36.4$\times$ longer than EAST's end-to-end training. EAST uses only 40GB of memory during training, which is highly efficient considering the ViT-G backbone alone requires 6GB for per-frame feature extraction. In contrast, prior work  requires 25TB to extract I3D features for a 10-minute video. Our efficiency stems from segmentation-by-detection enabling low frame-sampling rates (1-6 fps) versus high-FPS SOTA methods. For inference on a 10-minute 50Salads video, EAST uses only 4.3K GFLOPs—lower than all prior work when combining their feature extraction and model GFLOPs (Tab~\ref{tab:parameter}). While EAST has 152.58M parameters (mainly ViT-G), the added components are proportionally smaller than SOTA relative to their backbones (e.g., I3D's 24.57M parameters).

\begin{table}
	\centering
	\resizebox{\columnwidth}{!}{
		\begin{tabular}{c|cccc|c}
			\hline
			Method&MSTCN&ASFormer&LTContext&FACT&EAST\cr
			\hline
			Time per Epoch (min)&25.0&99.0&4.5&48.9&23.3\cr
			Number of Epochs&50&120&150&150&30\cr
			Training Duration (h)&20.8&198.0&11.3&122.3&11.6\cr
			Peak GPU Memory (MiB)&2464&5014&7358&18990&40804\cr
			\hline
		\end{tabular}
	}
	\caption{Comparison of training efficiency on Breakfast in terms of per-epoch time, total training time, total epochs, and memory usage on an H100 GPU. Our EAST matches MSTCN's per-epoch time, and reduces overall training time due to faster convergence of learning and capability to work with downsampled videos. Higher memory usage of EAST is due to taking the RGB input rather than pre-computed features.}
	\label{tab:abl_time}
\end{table}

\subsection{CEA Adapter and Backbones}
\label{sec:abl_detector}

This section evaluates the proposed CEA adapter for fine-tuning the backbone alongside different backbone networks. We analyze their impact on EAST's detector performance, as the accuracy of its output -- i.e., action proposals -- is critical for the effectiveness of the subsequent aggregation and refinement modules. To assess this, we report average precision across varying IoU thresholds and mAP.

\begin{table}[t]
\centering
\resizebox{0.9\columnwidth}{!}{
	\begin{tabular}{c|c|ccccc|c}
			\hline
			Dataset&Method&0.3&0.4&0.5&0.6&0.7&mAP\cr	
			\hline
			\multirow{4}{*}{GTEA}& No adapter&89.4&87.1&83.0&74.0&66.1&79.9\cr	
            \cline{2-8}
            &Standard \cite{houlsby2019parameter}&94.9&92.9&90.3&83.4&75.6&87.4\cr	
			&TIA \cite{liu2024end}&94.7&92.9&88.2&83.1&76.4&87.1\cr	
               &CEA&{\bf 95.2}&{\bf 93.8}&{\bf 91.1}&{\bf 84.9}&{\bf 77.9}&{\bf 88.6}\cr	
			\hline
			\multirow{4}{*}{Breakfast}	& No adapter&64.7&61.9&58.3&52.0&42.8&55.9\cr	
            \cline{2-8}
            &Standard \cite{houlsby2019parameter}&73.8&71.3&68.0&62.2&53.9&65.9\cr	
			&TIA \cite{liu2024end}&74.9&72.5&69.2&63.7&54.8&67.0\cr	
            	&CEA&{\bf 75.8}&{\bf 73.2}&{\bf 70.1}&{\bf 63.9}&{\bf 55.7}&{\bf 67.7}\cr	
                                \hline
		\end{tabular}
}
\caption{Average precision at varying IoUs and mAP of EAST's detector  on Breakfast for different adapters used in fine-tuning of the ViT-G backbone. The row denoted with "No adapter" reports the results when ViT-G is not fine-tuned, i.e., when EAST uses precomputed MAEv2 frame features of the frozen ViT-G pretrained with VideoMAEv2.}
\label{tab:detector_mAP}
\end{table}

\begin{figure}[t]
    \centering
    \includegraphics[width=0.8\columnwidth]{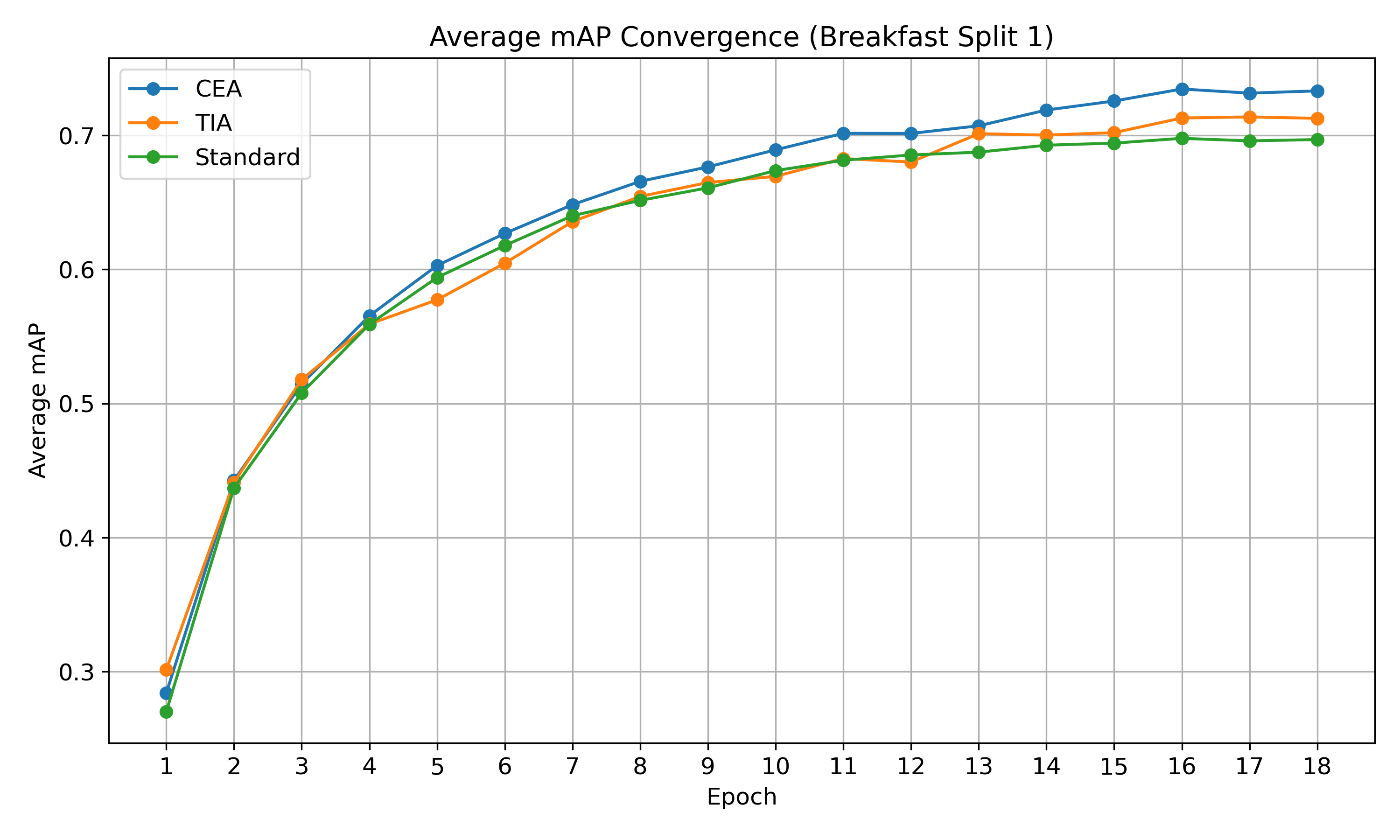}
    \caption{The plots of EAST's detector mAP vs. training epochs on Breakfast, for different adapters placed in the backbone, including the Standard Adapter\cite{houlsby2019parameter}, TIA \cite{liu2024end}, and our proposed CEA. The CEA adapter consistently achieves higher mAP across all epochs, and exhibits a faster convergence rate compared to the Standard and TIA adapters.}
    \label{fig:converge}
    \vspace{-10pt}
\end{figure}

{\bf Adapter.} Table~\ref{tab:detector_mAP} compares the performance of EAST's detector on GTEA and Breakfast when using different adapters for fine-tuning the ViT-G backbone, including the Standard Adapter \cite{houlsby2019parameter}, TIA \cite{liu2024end}, and our proposed CEA. We also conduct an ablation study where ViT remains frozen during training, without any adapter-based fine-tuning. In this setting, EAST effectively operates with precomputed MAEv2 frame features extracted from the ViT-G backbone pretrained with VideoMAEv2 \cite{wang2023videomae}.
Our CEA gives the best performance across all metrics. Regarding computational efficiency, Table~\ref{tab:adapter_gflops} shows that our CEA requires only 0.04 more GFLOPs than the Standard Adapter, whereas TIA incurs an additional 5.8 GFLOPs, demonstrating CEA's significantly lower computational cost compared to TIA.
Fig.~\ref{fig:converge} presents the  plots of mAP vs. training epochs on Breakfast, for different adapters placed in the backbone, showing that CEA consistently achieves the best mAP in every epoch.

\begin{table}
	\centering
	\resizebox{0.7\columnwidth}{!}{
	\begin{tabular}{c|cc|c}
		\hline
		Methods &Standard \cite{houlsby2019parameter} &TIA \cite{liu2024end}&CEA\cr
			\hline
GFLOPs &46.06 & 51.86 & 46.10\cr
		\hline
	\end{tabular}
	}
\caption{Computational cost (GFLOPs) of different adapters used for fine-tuning the backbone on a video sequence of 768 frames.}
\label{tab:adapter_gflops}
\end{table}

\begin{table}[t]
		\centering
			\resizebox{\columnwidth}{!}{
		\begin{tabular}{c|cc|ccccc|c}
			\hline
			Backbone&Dim &Mem &0.3&0.4&0.5&0.6&0.7&mAP\cr
			\hline
           
                ViT-S&384&3787&66.4&63.7&58.7&53.3&43.9&57.2\cr
                ViT-B&768&9434&68.9&66.6&62.8&55.3&47.3&60.2\cr
                ViT-L&1024&16920&73.1&69.4&65.6&58.3&49.3&63.1\cr
                ViT-G&1408&40804&78.8&76.3&72.4&65.0&56.7&69.8\cr
			\hline
		\end{tabular}}
  \caption{Evaluation of  EAST's detector on Breakfast when using different backbone networks. The table reports the backbone's output feature dimension (Dim), GPU memory usage (Mem, in MiB) during training, average precision at varying IoUs, and mAP.}
  \label{tab:backbone_ablation}
	\end{table}
 
\begin{table}[ht]
\centering
\resizebox{0.95\columnwidth}{!}{

\begin{tabular}{l|l|ccccc}
\hline
Dataset & Method & F1@10 & F1@25 & F1@50 & Edit & Acc \\
\hline
\multirow{4}{*}{GTEA} 
& Baseline & 87.3 & 84.7 & 75.9 & 82.5 & 86.3 \\
& MSTCN & 93.2 & 93.0 & 88.7 & 92.1 & 83.8 \\
& ASFormer & 86.8 & 84.0 & 74.3 & 83.1 & 83.6 \\
& Proposal Aug  & 95.8 & 95.4 & 91.7 & 95.4 & 87.1 \\
\hline
\multirow{4}{*}{Breakfast}
& Baseline & 78.2 & 74.2 & 65.8 & 75.1 & 77.6 \\
& MSTCN & 85.6 & 81.3 & 71.5 & 83.4 & 80.3 \\
& ASFormer & 84.1 & 79.9 & 69.7 & 82.2 & 80.2 \\
& Proposal Aug & 86.2 & 82.2 & 71.8 & 84.5 & 82.8 \\
\hline
\end{tabular}
}
\caption{Action segmentation results using identical action proposals from our detector, comparing refinement strategies. “Baseline” denotes framewise classification via Eq.~\ref{eq:aggregation} without augmentation (Sec.~\ref{sec:augmentation}). “MSTCN” and “ASFormer” apply their respective refinement networks to our generated proposals. “Proposal Aug” (EAST) incorporates our augmentation during training for improved refinement learning. The comparison between Baseline and Proposal Aug isolates our proposed augmentation' impact.}
\label{tab:detector_FEA}
\end{table}

{\bf Backbone.} Tab.~\ref{tab:backbone_ablation} compares EAST's detector performance on Breakfast when using different ViT backbones, from the smallest ViT-S to the largest ViT-G, pre-trained with VideoMAE \cite{tong2022videomae} and VideoMAEv2 \cite{wang2023videomae}. ViT-G is pre-trained on the unlabeled hybrid dataset \cite{wang2023videomae}, while the other ViT variants are pre-trained on the Kinetics-400 dataset \cite{zisserman2017kinetics}. Tab.~\ref{tab:backbone_ablation} shows that EAST with ViT-G outperforms the other alternatives across all metrics. Therefore, we select  ViT-G as our default backbone, and set that its output fine-tuned frame features $\mathcal{X}$ have a dimensionality of 1408. In contrast, previous work uses pre-computed I3D \cite{I3D} or TSM \cite{lin2019tsm} features with a dimensionality of 2048.

\subsection{Aggregator and Sensitivity to FPS}
\label{sec:abl_aggregator}
Tab.~\ref{tab:detector_FEA} reports the action segmentation performance of EAST’s aggregation and refinement modules on GTEA and Breakfast. It also includes ablations where we replace the aggregator and TCN module with the refinement stages of MS-TCN \cite{AbuFarha2019MSTCNMT} and ASFormer \cite{yi2021asformer}, or omit refinement of EAST’s detector output altogether.
 The results demonstrate that EAST's aggregation and refinement gives the best segmentation performance across all metrics on both datasets.

\begin{table}[t]
		\centering
	\resizebox{0.7\columnwidth}{!}{
		\begin{tabular}{c|ccccc}
			\hline
			FPS &\multicolumn{3}{c}{F1@\{10,25,50\}}&Edit&Acc\cr
			\hline
            1&84.1 & 79.8 & 69.6 & 81.7 & 80.4\cr
                3&86.2 & 82.2 & 71.8 & 84.5 & 82.8\cr
			\hline
		\end{tabular}
  }
  \caption{Sensitivity of EAST performance to the frame sampling rate (FPS)  on Breakfast.}
  \label{tab:fps}
\end{table}

Tab.~\ref{tab:fps} evaluates EAST's sensitivity to the video downsampling rate on Breakfast. EAST maintains SOTA performance at low FPS rates and improves as the frame rate increases, subject to memory and compute constraints.

\begin{table}[t]
\centering
\resizebox{\columnwidth}{!}{
	\begin{tabular}{c|c|ccccc}
			\hline
			Backbone&Method&\multicolumn{3}{c}{F1@\{10,25,50\}}&Edit&Acc\cr
			\hline	
			\multirow{7}{*}{Frozen}	&		MSTCN('CVPR19) \cite{AbuFarha2019MSTCNMT}&  85.8& 83.4 & 69.8 & 79.0 &76.3  \cr	
			&			ASFormer('BMCV21) \cite{yi2021asformer}&	90.1&88.8&79.2&	84.6&79.7  \cr	
			&	UVAST('ECCV22) \cite{uvast2022ECCV}&	92.7 &91.3 &81.0 &92.1 &80.2\cr
			&RTK('ICCV23)	\cite{jiang2023video}&91.2 &90.6 &83.4& 87.9 &80.3  \cr	
			&DiffAct ('ICCV23) \cite{liu2023diffusion}& 		92.5&91.5	&84.7&	89.6 & 82.2  \cr
			&FACT ('CVPR24 \cite{lu2024fact})&  93.5 &92.1& 84.1 &91.4 & 86.1 \cr
            &BaFormer ('NIPS24 \cite{wang2024efficient})& 92.0 & 91.3 & 83.5 & 88.7 & 83.0\cr
			\hline
                ViT-G fine-tuned    &EAST& \textbf{95.8} & \textbf{95.4} & \textbf{91.7} & \textbf{95.4} & \textbf{87.1}\cr
			
			\hline
		\end{tabular}
}
\caption{Action segmentation on GTEA.}
\label{tab:performance_comparison_gtea}
\end{table}

\subsection{Comparison with SOTA\label{Sec4.2}}
Tables~\ref{tab:performance_comparison_gtea}--\ref{tab:performance_comparison_assembly101} compare EAST  with SOTA on four datasets. EAST consistently outperforms prior work across all datasets, achieving substantial improvements in all metrics. On Assembly101, EAST surpasses previous methods by 7.2 points in accuracy and 9.5 points in Edit score. Furthermore, EAST demonstrates F1@{50} score gains of 7.0, 3.6, 5.2, and 10.2 percentage points on the GTEA, 50Salads, Breakfast, and Assembly101 datasets, respectively.
\begin{table}[t]
\centering
\resizebox{\columnwidth}{!}{
	\begin{tabular}{c|c|ccccc}
			\hline
			Backbone&Method&\multicolumn{3}{c}{F1@\{10,25,50\}}&Edit&Acc\cr
			\hline	
			\multirow{7}{*}{Frozen}	&		MSTCN('CVPR19) \cite{AbuFarha2019MSTCNMT}&  76.3& 74.0& 64.5 & 67.9 &80.7  \cr	
			&		ASFormer('BMCV21) \cite{yi2021asformer} &		85.1&	83.4&76.0&	79.6		&85.6	\cr	
			&	UVAST('ECCV22)\cite{uvast2022ECCV} &89.1& 87.6 &81.7 &83.9 &87.4\cr
			&		RTK('ICCV23) \cite{jiang2023video}&	87.4 &86.1 &79.5& 81.4 &85.9\cr
			&		LTContext('ICCV23) \cite{bahrami2023much}&	89.4& 87.7&82.0&83.2&	87.7 \cr	
			&		DiffAct ('ICCV23) \cite{liu2023diffusion}& 	90.1&89.2&83.7&  	85.0&88.9    \cr	
             &BaFormer ('NIPS24 \cite{wang2024efficient})&89.3 & 88.4 & 83.9 & 84.2 & 89.5\cr
			\hline  	
			ViT-G fine-tuned    &EAST& \textbf{93.1} & \textbf{91.9} & \textbf{88.6} & \textbf{88.4} & \textbf{91.9} \cr	
                
			\hline
		\end{tabular}
}
\caption{Action segmentation on 50salads.}
\label{tab:performance_comparison_50salads}
\end{table}

\begin{table}[t]
\centering
\resizebox{\columnwidth}{!}{
	\begin{tabular}{c|c|ccccc}
			\hline
			Backbone&Method&\multicolumn{3}{c}{F1@\{10,25,50\}}&Edit&Acc\cr
			\hline	
			\multirow{8}{*}{Frozen}&	MSTCN('CVPR19) \cite{AbuFarha2019MSTCNMT}&	52.6&	48.1&37.9&	61.7	&66.3	\cr	
			&	ASFormer('BMCV21) \cite{yi2021asformer}&		76.0&70.6&57.4&		75.0&73.5	\cr	
			&	UVAST('ECCV22)\cite{uvast2022ECCV} &	75.9& 70.0& 57.2 &76.5& 66.0\cr
			&		RTK('ICCV23) \cite{jiang2023video}&	76.9 &72.4 &60.5& 76.1 &73.3\cr	
			&	LTContext('ICCV23) \cite{bahrami2023much}& 77.6 &72.6&60.1&77.0&74.2 \cr	
			&	DiffAct ('ICCV23) \cite{liu2023diffusion}& 		80.3&75.9&		64.6	&78.4	&76.4	  \cr
			& FACT ('CVPR24) \cite{lu2024fact} &81.4& 76.5& 66.2& 79.7 &76.2 \cr
            &BaFormer ('NIPS24 \cite{wang2024efficient})&79.2 & 74.9 & 63.2 & 77.3 & 76.6\cr
			\hline  
			ViT-G fine-tuned    & EAST&  \textbf{86.2} & \textbf{82.2} & \textbf{71.8} & \textbf{84.5} & \textbf{82.8}\cr
			\hline
		\end{tabular}
}
\caption{Action segmentation Breakfast.}
\label{tab:performance_comparison_breakfast}
\end{table}

\begin{table}[t]
\centering
\resizebox{\columnwidth}{!}{
	\begin{tabular}{c|c|ccccc}
			\hline
			Backbone&Method&\multicolumn{3}{c}{F1@\{10,25,50\}}&Edit&Acc\cr
			\hline	
			\multirow{5}{*}{Frozen}&	MS-TCN++ (PAMI'20) \cite{2020_PAMI_Li}        &  31.6 & 27.8 & 20.6 &  30.7 & 37.1	\cr	
			&	UVAST (ECCV'22) \cite{uvast2022ECCV}  & 32.1  & 28.3 & 20.8 &  31.5 & 37.4	\cr	
   	&	ASFormer ('BMCV21)  \cite{yi2021asformer}                      &  33.4  & 29.2 & 21.4 &  30.5 & 38.8      \cr	
			&	C2F-TCN (TPAMI'23) \cite{singhania2021coarse} & 33.3  & 29.0 & 21.3 & 32.4 & 39.2\cr	
			&	LTContext ('ICCV23) \cite{bahrami2023much}                             & 33.9 & 30.0& 22.6 & 30.4 & 41.2 \cr
   
			\hline 
			ViT-G fine-tuned   & EAST& \textbf{42.3} & \textbf{39.4} & \textbf{32.8} & \textbf{39.9} & \textbf{48.4}\cr	
			\hline
		\end{tabular}
}
\caption{Action segmentation on Assembly101.}
\label{tab:performance_comparison_assembly101}
\end{table}

While maintaining comparable per-epoch processing times to prior work, EAST significantly reduces the total number of training epochs required for convergence. On  Breakfast, EAST converges in 30 epochs, compared to 150 for FACT \cite{lu2024fact} and 1000 for DiffAct \cite{liu2023diffusion}. On Assembly101, EAST requires only 12 epochs, whereas LTContext \cite{bahrami2023much} and C2F-TCN \cite{singhania2021coarse} require 120 and 200 epochs. 

\subsection{Qualitative Results}
\label{sec:qualitative}
\begin{figure}
    \centering
    \includegraphics[width=\columnwidth]{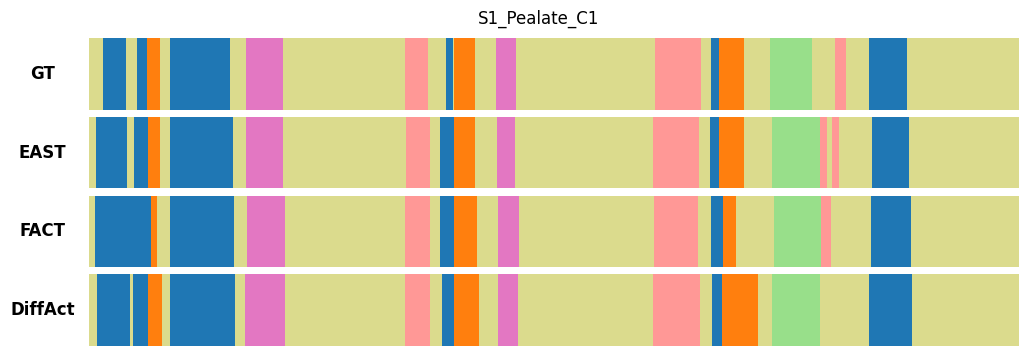}
    \includegraphics[width=\columnwidth]{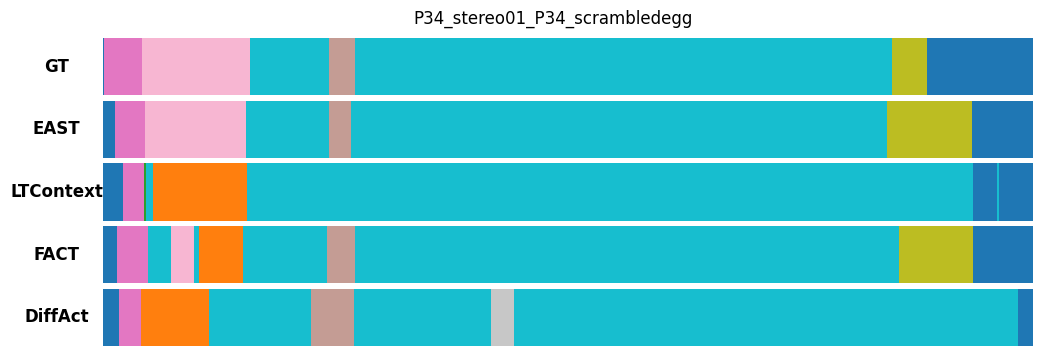}
    \includegraphics[width=\columnwidth]{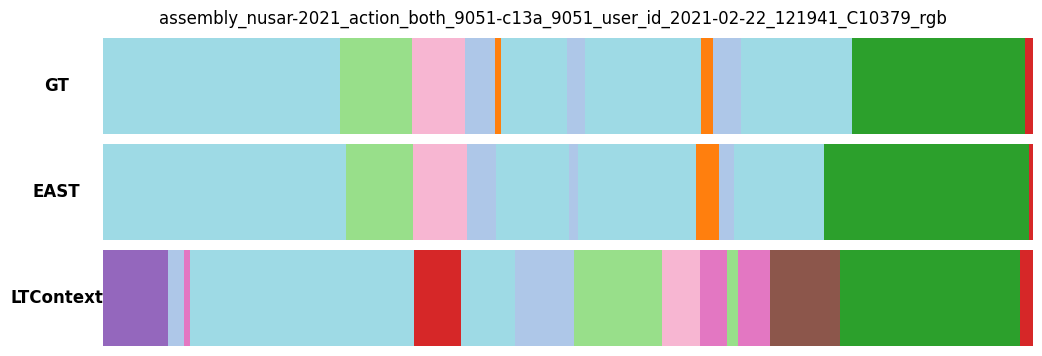}
    \caption{Segmentation results on sample videos from GTEA (top),  Breakfast (middle), and Assembly101 (bottom). For each video sequence, the top row shows the color-coded ground-truth action instances, middle row shows EAST output, and bottom row shows SOTA results generated using publicly available models. }
    \label{fig:SOTAs}
\end{figure}

Fig.~\ref{fig:SOTAs} compares of EAST segmentation results with the ground truth and SOTA on sample videos from GTEA, Breakfast, and Assembly101. In the top example video from GTEA, EAST successfully detects all actions, whereas the other methods miss at least one action. On the middle example video from Breakfast, the SOTA methods exhibit varying degrees of missed actions, spurious predictions, or oversegmentation. In contrast, EAST provides high-quality segmentation closely aligned with the ground truth. Finally, in the bottom video from Assembly101, EAST fails to detect a brief action instance and incorrectly identifies an action boundary too early -- both of these are ambiguous edge cases that are challenging to discern even through visual inspection.

\section{Conclusion}
We have introduced EAST -- the first fully end-to-end trainable action segmenter. EAST performs segmentation by detection, which enables temporal downsampling of input videos, significantly reducing computational costs. EAST takes RGB frames, sampled at a low frame rate, as input to a large-scale backbone. The backbone is fine-tuned using our Contract-Expand Adapter (CAE). CAE is especially effective in reducing the computational costs of end-to-end training by leveraging spatial pooling. The backbone features are passed to the detector to predict action proposals, which are then aggregated and refined to produce final framewise labeling at the original unsampled frame rate. We have also specified a novel proposal-based data augmentation that increases uncertainty of the detector's output during training, effectively simulating test-time conditions. EAST outperforms prior work across all metrics on the GTEA, 50Salads, Breakfast, and Assembly101 datasets, while maintaining comparable processing times, even with the additional step of extracting frame features by the backbone. We have conducted a comprehensive ablation study to evaluate EAST's performance under various configurations. Additional results are presented in the supplements.

\noindent\textbf{Acknowledgement}: This work has been supported by USDA NIFA award No.2021-67021-35344.\newpage

{
    \small
    \bibliographystyle{ieeenat_fullname}
    \bibliography{main}
}
\clearpage
\maketitlesupplementary

\section{Additional Results}
 {\bf Feature Extraction Time.}  As demonstrated in Table \ref{tab:abl_fea_extract}, offline feature extraction introduces significant time costs, ranging from hours (GTEA) to months (Assembly101). Therefore, our end-to-end method provides substantial reductions in both training and inference time compared to methods relying on pre-extracted features.

\noindent{\bf SOTA Methods with Low FPS Input.}  The impact of video downsampling on the SOTA methods and EAST is evaluated on Breakfast in Table \ref{tab:05_downs_bf}. 
The SOTA methods are trained using I3D frame features sampled at 1 FPS, with  their output framewise classification subsequently upscaled to the original 15 FPS, as per their reported evaluation setting. Table \ref{tab:05_downs_bf} shows significant performance degradation for all the SOTA methods when working with the low FPS at the input. As shown in Tables~\ref{tab:05_downs_bf} and \ref{tab:fps}, EAST maintains the best performance at low FPS rates and improves as the frame rate increases, subject to memory and compute constraints.

\begin{figure}[b]
    \centering
    \includegraphics[width=\columnwidth]{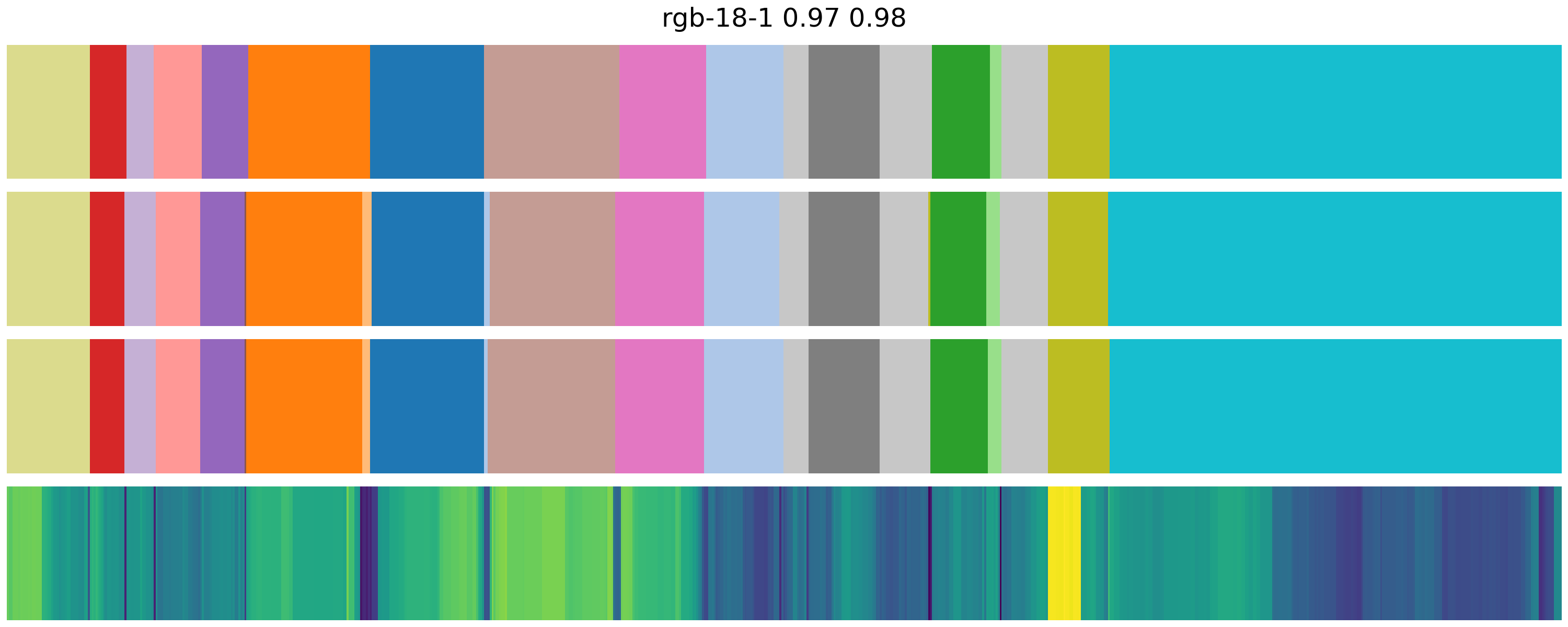}
    \includegraphics[width=\columnwidth]{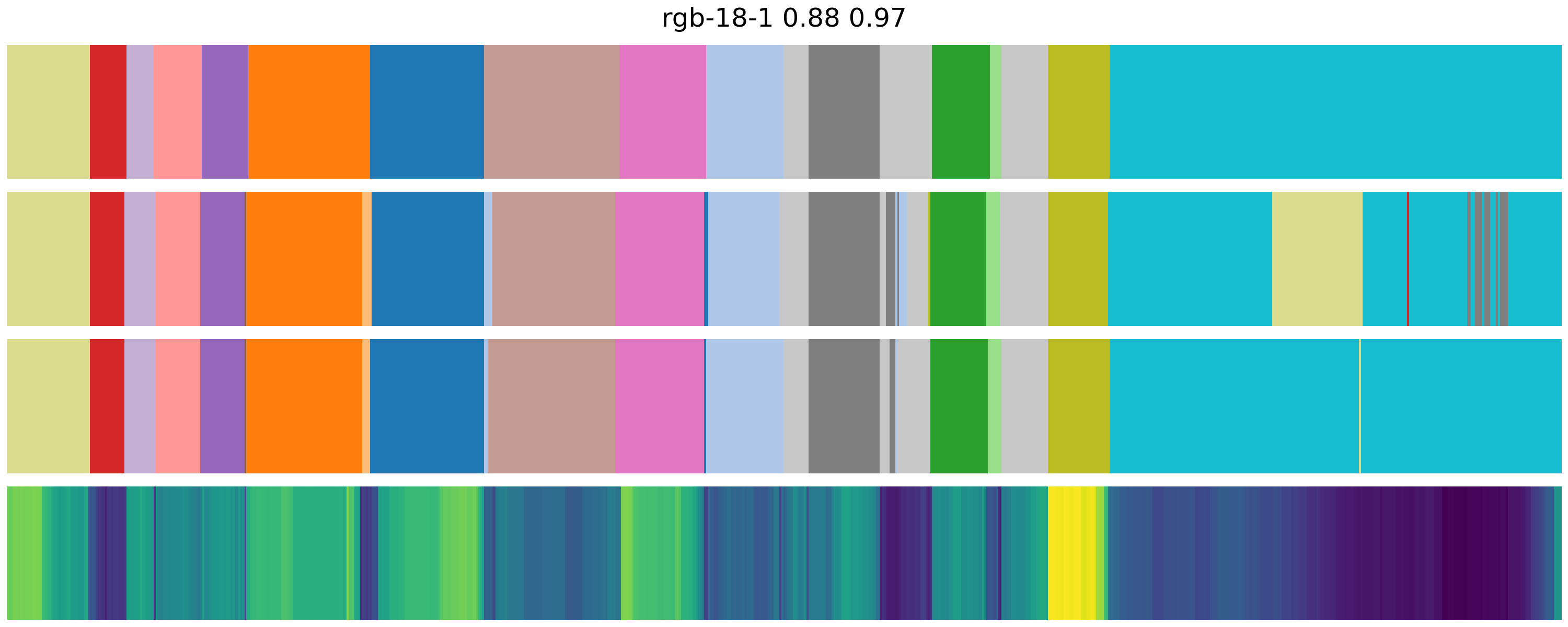}
    \includegraphics[width=\columnwidth]{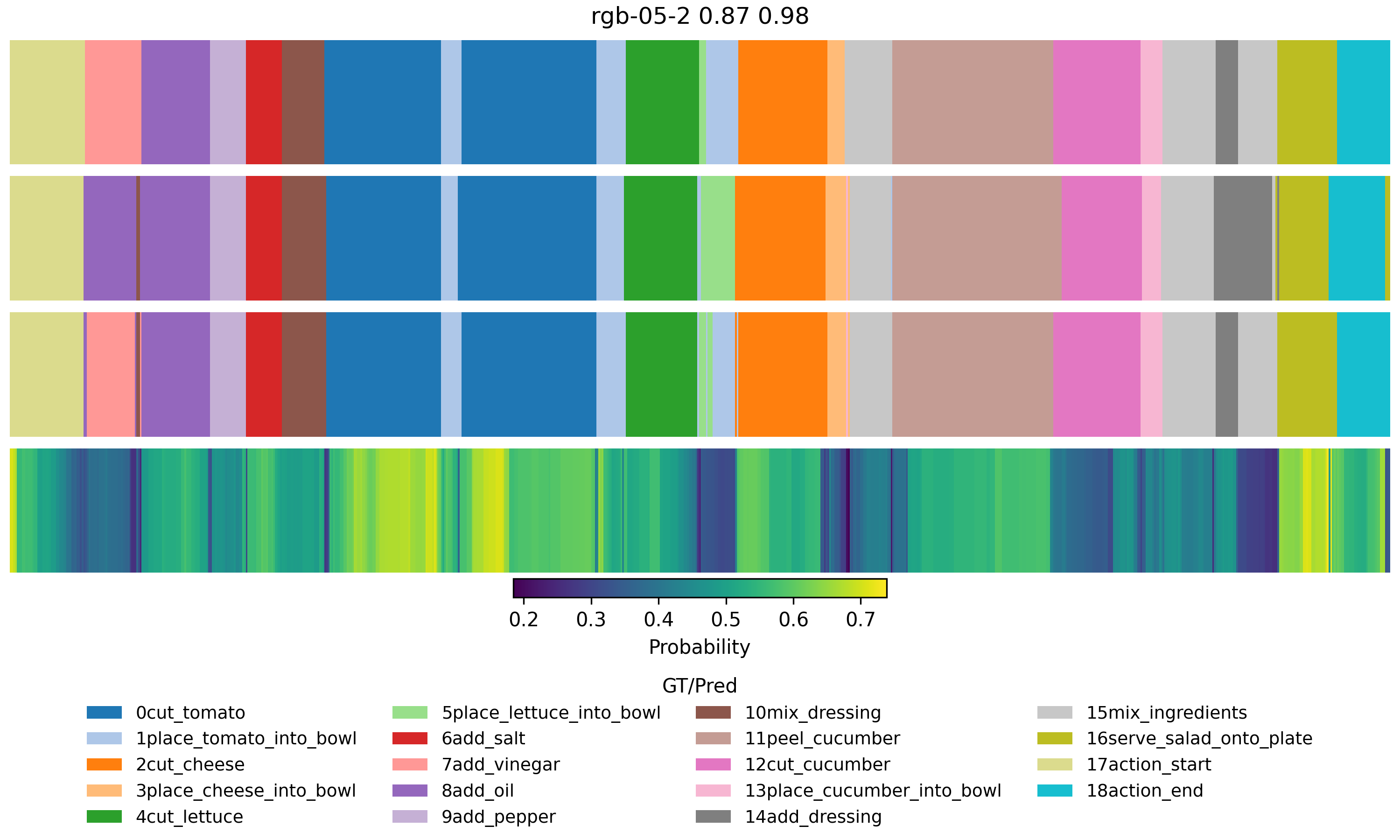}
    \caption{EAST detector's framewise classification and softmax scores on example videos from 50Salads. The top two videos depict the same training video (original vs. augmented predictions), while the bottom video, representing one of the worst cases, is from the evaluation set. A color-coded legend for frame labels and softmax score ranges is shown below.}
    \label{fig:50salads_uncert}
\end{figure}
\begin{figure}
    \centering
    \includegraphics[width=\columnwidth]{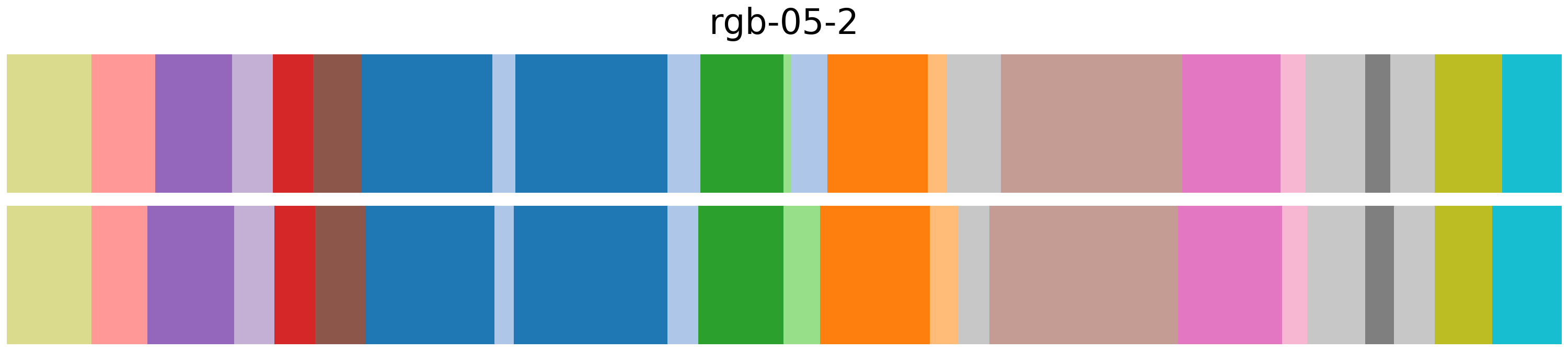}
    \caption{(top) Ground truth and (bottom) EAST's final framewise result for the video rgb-05-2 of 50Salads also considered  in Fig. \ref{fig:50salads_uncert}. }
    \label{fig:50salads_corr}
\end{figure}

\begin{table}[b]
	\centering
\resizebox{\columnwidth}{!}{
	\begin{tabular}{c|cccc}
		\hline
		Dataset&GTEA&50salads&Breakfast&Assembly101\cr
		\hline
		Avg. Time/Video (min)&5.6&57.8&10.5&63.7\cr
		Videos&28&50&1712&6108\cr
 Total Time (h) &2.6&48.2&300.3&6430.2\cr
		\hline
	\end{tabular}
    }
\caption{Feature extraction time for different datasets. ``Avg. Time/Video" shows the average processing time per video. ``Videos" denotes the number of videos. ``Total Time" indicates the overall extraction time; times are based on an H100 GPU.}
\label{tab:abl_fea_extract}
\end{table}

\begin{table}
	\centering
\resizebox{0.9\columnwidth}{!}{
	\begin{tabular}{c|c|ccccc}
		\hline
		Method&FPS&\multicolumn{3}{c}{F1@\{10,25,50\}}&Edit&Acc\cr
			\hline
				MSTCN \cite{AbuFarha2019MSTCNMT}&15&52.6&	48.1&37.9&	61.7	&66.3	\cr	
			('CVPR19) &1&72.5 & 65.8 & 49.8 & 71.1 & 67.9	\cr	
			\hline
			ASFormer \cite{yi2021asformer}&15&	76.0&70.6&57.4&		75.0&73.5	\cr	
			('BMCV21)&1 & 74.3 & 67.6 & 51.9 & 73.5 & 69.6\cr
				\hline
			{LTContext}  \cite{bahrami2023much}&15&77.6 &72.6&60.1&77.0&74.2\cr
				('ICCV23)&1			&77.2 & 70.5 & 56.1 & 74.1 & 69.8	\cr	
			\hline
{FACT} \cite{lu2024fact}&15&81.4& 76.5& 66.2& 79.7 &76.2 \cr
('CVPR24) 	&1&76.3 & 70.7 & 56.8 & 74.4 & 70.9\cr	
\hline
\multirow{2}{*}{EAST}&3& \textbf{86.2} & \textbf{82.2} & \textbf{71.8} & \textbf{84.5} & \textbf{82.8}\cr
&1&84.1 &  79.8 &   69.6 &  81.7 & 80.4 \cr

		\hline
	\end{tabular}
    }
\caption{Impact of FPS (frame per second) on SOTA and EAST performance on Breakfast.}
\label{tab:05_downs_bf}
\end{table}

\noindent {\bf Qualitative Results.} Fig.~\ref{fig:50salads_uncert} illustrates EAST detector's action segmentation on three example videos from 50Salads. For each video, frame labels and prediction scores are visualized in four rows, including (from top to bottom): (a) Ground-truth frame labels; (b) Predicted highest-scoring frame labels; (c) A hypothetical frame labeling guided by an oracle which replaces incorrect highest-scoring labels in (b) with the second-highest scoring class; and (d) maximum softmax score of the predicted class for each frame. The video title includes the video name and the accuracy of (b) and (c).  By comparing (a) and (b) in Fig.\ref{fig:50salads_uncert}, we observe that labeling errors predominantly occur in frames with low softmax scores, indicating that our detector is reliably trained. In (c), most errors can be corrected by replacing the incorrect highest-scoring class with the second highest, assuming access to an oracle. This highlights the potential for self-correction by refining predictions at frames with low softmax scores — the main purpose of the aggregator. Fig.\ref{fig:50salads_corr} demonstrates that EAST effectively refines the initially predicted action segments by the detector, enhancing their alignment with the ground-truth labels.


\end{document}